\documentclass[VANCOUVER,Times2COL]{WileyNJDv5} 

\articletype{Original Research}%

\newcommand{\defgr}{\mathrel{\mathop:\!\!=}}
\newcommand\norm[1]{\left\lVert#1\right\rVert}

\usepackage{lmodern}
\usepackage{subcaption}
\usepackage{makecell}

\begin{document}

\title{A Novel Distance-Based Metric for Quality Assessment in Image Segmentation}

\author[1]{Niklas Rottmayer}

\author[1]{Claudia Redenbach}

\authormark{Rottmayer and Redenbach}
\titlemark{A Novel Distance-Based Metric for Quality Assessment in Image Segmentation}

\address[1]{\orgdiv{Mathematics Department}, \orgname{RPTU University Kaiserslautern-Landau}, \orgaddress{\state{Rhineland-Palatinate}, \country{Germany}}}

\corres{Niklas Rottmayer \email{niklas.rottmayer@rptu.de}}

\fundingInfo{Federal Ministry of Education and Research (BMBF), Project: Synthetic Data for Machine Learning Segmentation of Highly Porous Structures from FIB-SEM Nano-tomographic Data (poSt), Funding number: 01IS21054A}

\abstract[Abstract]{The assessment of segmentation quality plays a fundamental role in the development, optimization, and comparison of segmentation methods which are used in a wide range of applications. With few exceptions, quality assessment is performed using traditional metrics, which are based on counting the number of erroneous pixels but do not capture the spatial distribution of errors. Established distance-based metrics such as the average Hausdorff distance are difficult to interpret and compare for different methods and datasets. In this paper, we introduce the Surface Consistency Coefficient (SCC), a novel distance-based quality metric that quantifies the spatial distribution of errors based on their proximity to the surface of the structure. Through a rigorous analysis using synthetic data and real segmentation results, we demonstrate the robustness and effectiveness of SCC in distinguishing errors near the surface from those further away. At the same time, SCC is easy to interpret and comparable across different structural contexts. }

\keywords{Segmentation, Quality Assessment, Distance-based, Synthetic Data}

\maketitle

\renewcommand\thefootnote{}
\renewcommand\thefootnote{\fnsymbol{footnote}}
\setcounter{footnote}{1}

\section{Introduction} 
Segmenting images into meaningful regions plays an important role in a wide range of applications including medical imaging, autonomous driving, and materials science. Accurate segmentation is often a prerequisite for downstream processing such as quantitative analysis or classification. In the supervised setting, quality assessment relies on the comparison of the segmentation result with a ground truth annotation. 
This assessment is essential when developing, improving, and comparing methods on different datasets, e.g., when training machine learning models. 

Quality metrics such as the Dice similarity coefficient (DSC) or the Hausdorff distance \cite{taha_metrics_2015} are frequently used in segmentation tasks and have proven useful across many domains. However, these metrics are not without limitations. For instance, DSC and intersection over union (IoU) summarize performance solely using ratios of correctly and incorrectly annotated pixels regardless of their position in the image. As many other metrics, they are sensitive towards class-imbalances. In contrast, the Hausdorff distance and its descendents collect spatial information from errors to assess quality. However, they yield unnormalized values because they scale with the number of errors. This limits the interpretability of individual values and complicates comparisons between different datasets and segmentations. In general, all quality metrics have certain limitations due to their reduction of complex images to a scalar \cite{reinke_understanding_2024}. Therefore, a precise assessment of performance requires the use of multiple quality metrics at once.

The literature related to the analysis, comparison and application of quality metrics in segmentation tasks is extensive, with several studies and surveys comparing multiple metrics against each other on various types of datasets. Wang et al. \cite{wang_image_2020} provide an overview of available quality metrics for supervised and unsupervised tasks and discuss the corresponding literature. Reinke et al. \cite{reinke_understanding_2024} offer a detailed list of quality metrics, their limitations and pitfalls which have to be considered in applications. Other articles such as \cite{pont-tuset_measures_2013,mageswari_analysis_2014,taha_metrics_2015,dey_social_2018} analyze and compare selected quality metrics on various types of image data in detailed studies. They primarily cover traditional metrics which can be expressed through the basic cardinalities true positives (TP), false positives (FP), false negatives (FN) and true negatives (TN). Most importantly, they do not consider the position of pixels and their geometric relations. Chicco and coauthors published multiple articles showing that the Matthews correlation coefficient (MCC) is superior to other traditional metrics for general quality assessment \cite{chicco_advantages_2020,chicco_matthews_2021,chicco_matthews_2021-1,chicco_matthews_2023}. Distance-based metrics often fall short in the discussion about quality and only contain a hand-full of metrics which are derived by weighing incorrectly annotated pixels by their distance to the structure surface. This results in unnormalized values which highly depend on the geometry and are therefore difficult to interpret without additional information. Several attempts were made to improve the robustness, reliability and interpretability of the Hausdorff distance such as averaging distances, using quantiles \cite{taha_metrics_2015}, adjusting the normalization \cite{aydin_usage_2021} or restricting the comparison only to the boundaries \cite{garcia-lamont_segmentation_2018}. However, none of the approaches was able to completely resolve all the issues at once. Nevertheless, distance-based quality assessment provides fundamental insight into the spatial distribution of errors and can benefit quality assessment. In particular, it allows to determine if predictions match the general shape of the ground truth and differ only in inconsistencies close to or far away from its surface. The importance of spatial information for segmentation tasks was also shown by Fend et al. \cite{fend_reconstruction_2021} who saw performance of AI segmentation models drastically increase when using a distance-based loss function. 

In this paper, we introduce a novel quality metric called the Surface Consistency Coefficient (SCC) which solves problems of previous distance-based metrics. It is normalized, quantifies performance based on spatial information, and is easy to interpret and compare across different data. In addition, SCC remains independent of the rate of errors. Instead of trying to assess quality alone, SCC should be combined with a traditional metric such as accuracy to assess both the quantity of errors and their geometric distribution independently of each other. This reduces the use of redundant information and enables a more precise assessment of quality and comparison of segmentations. To achieve this, SCC summarizes the geometric distribution of errors into a scalar value which differentiates between clustering of errors near or far from the geometric boundary. We rigorously analyze and validate SCC using a diverse set of geometries combined with synthetic segmentation results that are constructed by introducing systematic errors. Through comparison with established quality metrics such as DSC, MCC and the directed average Hausdorff distance (AHD), we show the utility of SCC for assessing quality and comparing available segmentations. 

\section{Methods}
We consider a binary segmentation or annotation of an image with pixels $X\subset \mathbb{Z}^d$ by the partition $S=\{S^0,S^1\}$, i.e., $S^{0} \cap S^{1}=\emptyset$ and $S^{0}\cup S^{1} = X$. The sets $S^{0}$ and $S^{1}$ are the back- and foreground, respectively. Let $S_{\textrm{gt}}$ denote the ground truth and $S_{\textrm{pr}}$ be a predicted annotation. Then, the set of incorrectly labeled pixels is $E\defgr (S^0_{\textrm{pr}} \cap S^1_{\textrm{gt}}) \cup (S^0_{\textrm{gt}} \cap S^1_{\textrm{pr}})$. This set can be divided to form the basic cardinalities
\begin{align*}
	\textrm{TP} &= \vert S^1_{\textrm{gt}} \cap S^1_{\textrm{pr}}\vert,
	&\textrm{FN} &=\vert S^1_{\textrm{gt}} \cap S^0_{\textrm{pr}}\vert, \\
	\textrm{FP} &= \vert S^0_{\textrm{gt}} \cap S^1_{\textrm{pr}}\vert,
	&\textrm{TN} &= \vert S^0_{\textrm{gt}} \cap S^0_{\textrm{pr}}\vert \label{eq:cardinalities}
\end{align*}
which are sufficient to express traditional metrics such as the Dice coefficient
\begin{equation}
    \textrm{DSC} = \frac{2\,\textrm{TP}}{2\,\textrm{TP}+\textrm{FP}+\textrm{FN}} \label{eq:DSC}
\end{equation}
and the Matthews correlation coefficient
\begin{equation}
    \textrm{MCC} = \frac{\textrm{TP} \cdot \textrm{TN} - \textrm{FP} \cdot \textrm{FN}}{\sqrt{(\textrm{TP}+\textrm{FP})(\textrm{TP} + \textrm{FN})(\textrm{TN} + \textrm{FP})(\textrm{TN} + \textrm{FN})}}. \label{eq:MCC}
\end{equation}
 We define the distance
\begin{equation}
    d_{\partial\textrm{gt}}(x) = \begin{cases}
        \min_{y\in S^1_{\textrm{gt}}} \norm{x - y} & \text{if } x\in S^0_{\textrm{gt}}\\
        \min_{y\in S^0_{\textrm{gt}}} \norm{x - y} & \text{if } x\in S^1_{\textrm{gt}}
        \end{cases}
\end{equation}
of any pixel $x\in X$ to the surface of the ground truth structure. Then, the directed average Hausdorff distance \cite{taha_metrics_2015} is defined by
\begin{equation}
    \textrm{AHD}(S_{\textrm{pr}},S_{\textrm{gt}}) \defgr \frac{1}{\vert X \vert}\sum_{x\in E} d_{\partial\textrm{gt}}(x).
\end{equation}
By construction, AHD is not normalized and scales with error quantity and distance. To separate number and distance of errors, we propose the surface consistency coefficient
\begin{equation}
    \textrm{SCC}(S_{\textrm{pr}},S_{\textrm{gt}}) \defgr \frac{1}{\vert E\vert}\sum_{x\in E} f(d_{\partial\textrm{gt}}(x)),
\end{equation}
with weighting function $f(r)\in[0,1]$ which controls the contribution of specific distances towards the quality assessment, i.e., incorrectly labeled pixels do not contribute directly with their distance but based on it. In addition, normalization by the number of errors $\vert E\vert$ removes scaling with error quantity. Together, these two changes result in a normalization of SCC to the interval $[0,1]$. 

\begin{figure}[ht]
    \centering
    \begin{subfigure}{0.24\linewidth}
        \includegraphics[width=\linewidth]{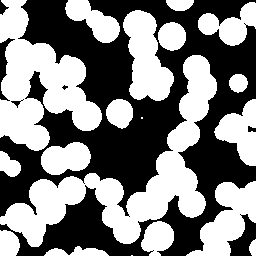}
        \subcaption{Structure}
    \end{subfigure}
    \begin{subfigure}{0.24\linewidth}
        \includegraphics[width=\linewidth]{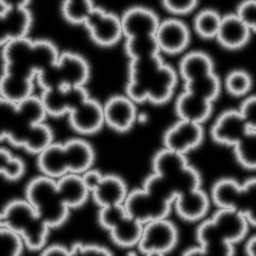}
        \subcaption{$a=0.5, k=3$}
    \end{subfigure}
    \begin{subfigure}{0.24\linewidth}
        \includegraphics[width=\linewidth]{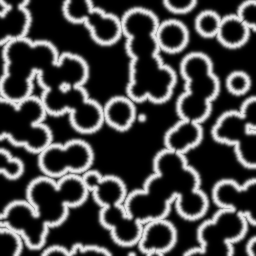}
        \subcaption{$a=1, k=3$}
    \end{subfigure}
    \begin{subfigure}{0.24\linewidth}
        \includegraphics[width=\linewidth]{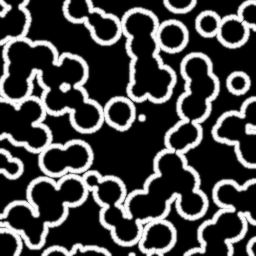}
        \subcaption{$a=2, k=3$}
    \end{subfigure}\\
    \begin{subfigure}{0.24\linewidth}
        \includegraphics[width=\linewidth]{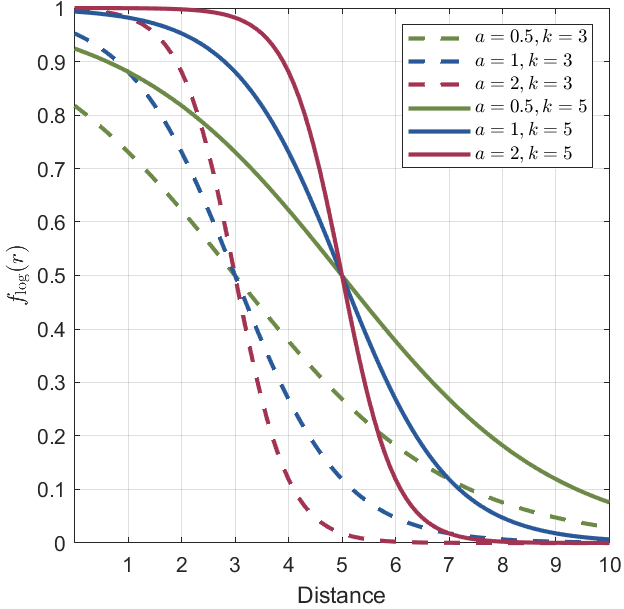}
        \subcaption{Weight functions}
    \end{subfigure}
    \begin{subfigure}{0.24\linewidth}
        \includegraphics[width=\linewidth]{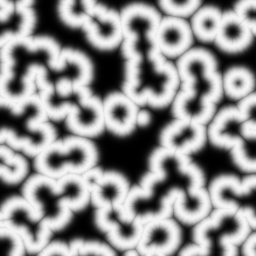}
        \subcaption{$a=0.5, k=5$}
    \end{subfigure}
    \begin{subfigure}{0.24\linewidth}
        \includegraphics[width=\linewidth]{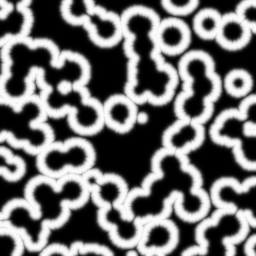}
        \subcaption{$a=1, k=5$}
    \end{subfigure}
    \begin{subfigure}{0.24\linewidth}
        \includegraphics[width=\linewidth]{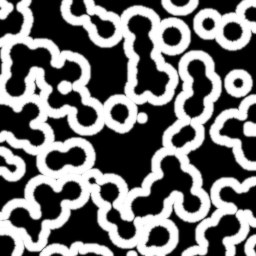}
        \subcaption{$a=2, k=5$}
    \end{subfigure}
    \caption{(a) Data:  section image of a 3D volume of overlapping spheres. (b-d,f-h) Weight maps using different parameter combinations for $f_{\log}$ (e). $a=1, k=5$ defines a suitable proximity range and transition speed for this geometry.}
    \label{fig:WeightMap}
\end{figure}

As weighting function, we propose the logistic function
\begin{equation}
    f_{\textrm{log}}(r) = \frac{1}{1+\exp\left(-a( r - k)\right)} \label{eq:LogisticFunction}
\end{equation}
with scaling parameter $a>0$ and shift $k\geq 0$ which we indicate by writing $\textrm{SCC}_{a,k}$. As illustrated in Figure \ref{fig:WeightMap}, pixels near the structure's surface have larger weights and pixels far from it are almost ignored. This can be interpreted as pixels becoming increasingly less important when focusing on quality near the surface. The values $k$ and $a$ are adjustable and should be chosen to reflect the desired understanding of proximity. We call $k$ the proximity range since it marks the distance from the surface at which pixels transition from being considered still close to rather distant indicated by $f_{\textrm{log}}(k) = 0.5$. The parameter $a$ controls the speed of this transition. We suggest to choose $a\approx4/(k_{\max}-k)$ where $k_{\max}\geq k$ is the smallest distance from the surface at which an error is without a doubt far away. The quantity $2(k_{\max}-k)$ is the transition width over which the function falls from approximately 1 to close to 0 and reflects the range over which errors become increasingly less proximal. In the limit $a\rightarrow\infty$, $f_{\textrm{log}}$ converges to a step function, i.e. a direct transition from proximal to distant. In general, the value of SCC relates roughly to the fraction of errors found outside the proximity range. Hence, small values indicate that errors are predominantly found near the surface, while large values imply that errors appear mostly outside the proximity range.

\section{Results and Discussion}
\subsection{Study of Synthetic Geometries}
In the following, we perform a rigorous analysis of SCC on a variety of different geometric structures to validate its use in quality assessment and highlight several key properties that set it apart from other quality metrics. For this, we consider 20 different 3D geometries derived from stochastic models with an image size of 512$^3$ pixels. The geometries differ in their particle shape and (volume) density to provide a diverse set of conditions. More precisely, we considered five different shapes (spheres, cubes, cylinders, ellipsoids and cuboids) and four different volume densities (10\%, 30\%, 50\% and 70\%) and generated one realization for each pair. The particles were rotated uniformly to form isotropic systems. For volume density 10\%, the particles were placed in the image such that they do not overlap. In all other cases, realizations were drawn from Boolean models \cite{chiu_stochastic_2013} for which the number of particles is Poisson distributed and the particle positions are independent and uniform in the image. Hence, particles are allowed to overlap and form complex geometries. The size of particles is constant and calculated such that each particle shape has the same ratio of surface area to volume, see Table \ref{tab:ParticleSize} for values. This results in the same expected surface area for the structures with the same volume density.

\begin{table}[ht]
    \caption{Parameters of each particle shape used for generating the 3D geometries. The ratio of surface area to volume for each shape is 0.2.}\label{tab:ParticleSize}
    \centering
    \begin{tabular}{c c}\hline
        \textbf{Shape} & \textbf{Parameters} [px]\\\hline
        Sphere & $radius = 15$ \\
        Cube & $edge~length = 30$ \\
        Cylinder & $radius=10.5$, $height=210$ \\
        Ellipsoid & $semiaxes = (8.46,25.39,84.63)$ \\
        Cuboid & $edge~lengths = (14.33,43,143.33)$ \\\hline
    \end{tabular}
    
\end{table}

\begin{figure}[h]
    \centering
    \begin{subfigure}{0.24\linewidth}
        \includegraphics[width=\linewidth]{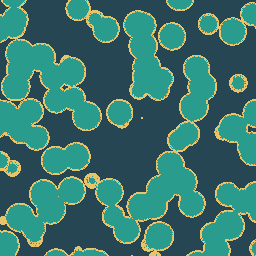}
        \subcaption{Erosion}
    \end{subfigure}
    \begin{subfigure}{0.24\linewidth}
        \includegraphics[width=\linewidth]{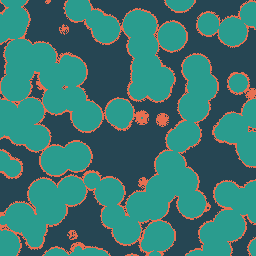}
        \subcaption{Dilation}
    \end{subfigure}
    \begin{subfigure}{0.24\linewidth}
        \includegraphics[width=\linewidth]{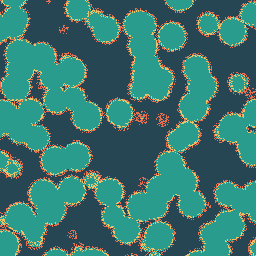}
        \subcaption{Fuzzy Edge}
    \end{subfigure}
    \begin{subfigure}{0.24\linewidth}
        \includegraphics[width=\linewidth]{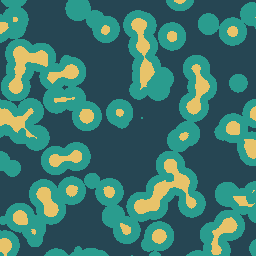}
        \subcaption{FN Cluster}
    \end{subfigure}
    \begin{subfigure}{0.24\linewidth}
        \includegraphics[width=\linewidth]{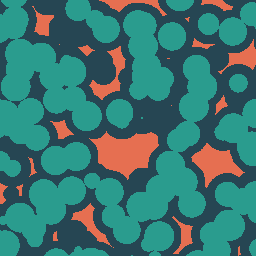}
        \subcaption{FP Cluster}
    \end{subfigure}
    \begin{subfigure}{0.24\linewidth}
        \includegraphics[width=\linewidth]{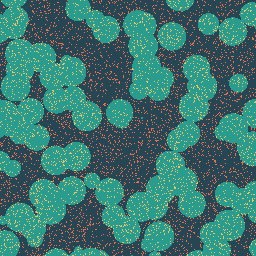}
        \subcaption{Uniform}
    \end{subfigure}
    \begin{subfigure}{0.24\linewidth}
        \includegraphics[width=\linewidth]{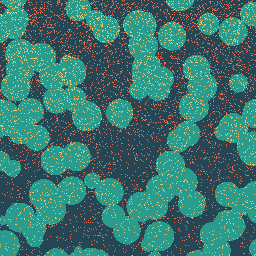}
        \subcaption{Nonuniform}
    \end{subfigure}
    \begin{subfigure}{0.24\linewidth}
        \includegraphics[width=\linewidth]{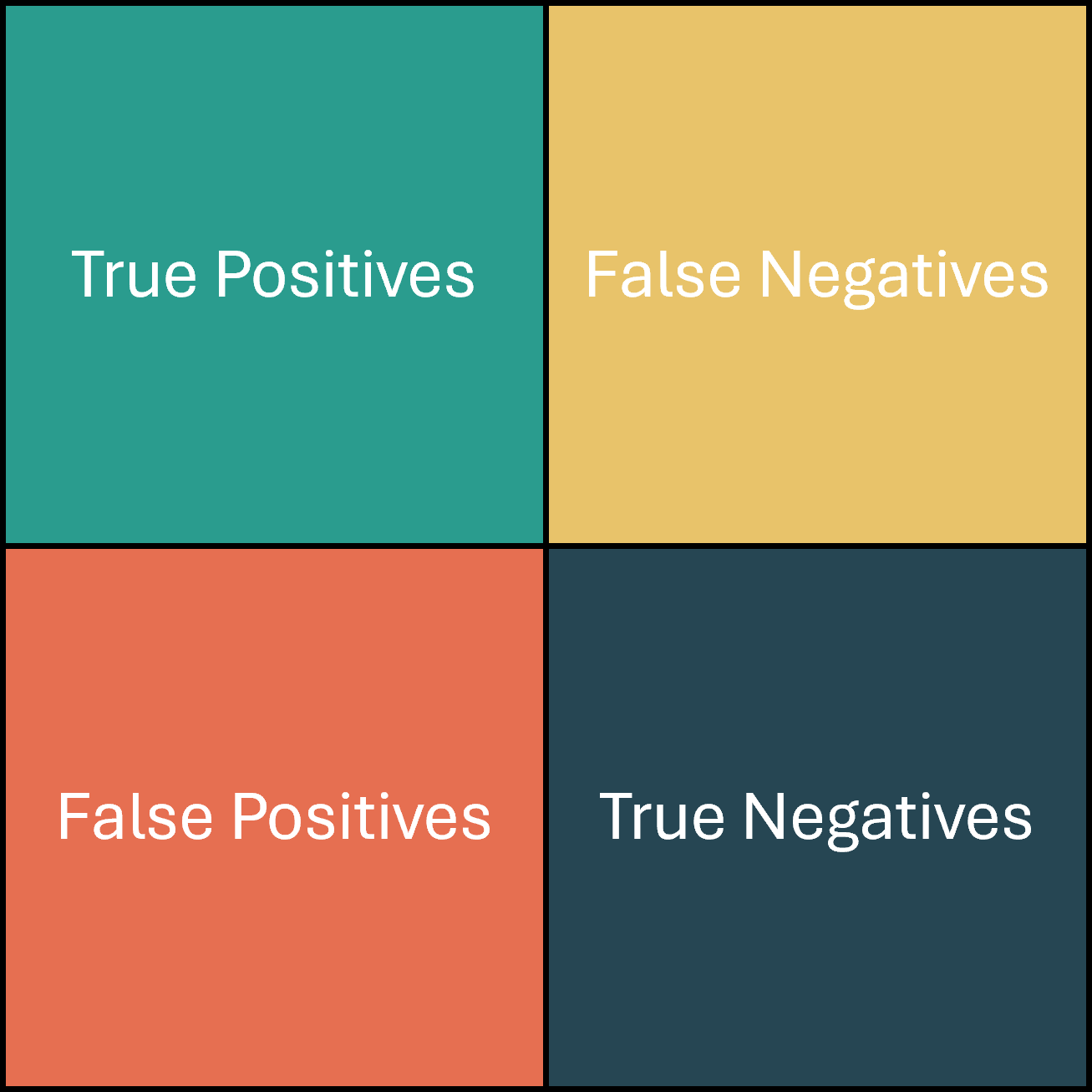}
        \subcaption{Color Legend}
    \end{subfigure}
    \caption{Synthetic segmentations created by introducing systematic errors with 15\% error rate visualized on a 256$^2$ section image from a 3D volume of overlapping spheres.}
    \label{fig:SyntheticSegmentations}
\end{figure}

To analyze the influence of distances and error type on SCC and other quality metrics, we generate synthetic segmentation results from each geometry by introducing systematic errors such as over-segmentation, under-segmentation or randomness in different parts of the image. This is achieved using morphological operations and the Euclidean distance transform (EDT), as illustrated in Figure \ref{fig:SyntheticSegmentations} and Table \ref{tab:SyntheticSegmentations}. 

We achieved exact error rates when using morphological operations by uniformly mislabeling pixels closest to the edge. Using erosion as an example, we first erode the structure by the largest ball which yields less errors than desired. Afterwards, foreground pixels adjacent to the newly formed surface are uniformly selected and mislabeled until the error rate is reached. By varying the error rate between 1\%-15\% we also study the influence of error quantity. In combination, this allows us to draw evidence-based conclusions on the general utility of SCC and other metrics. We compare SCC to the traditional metrics DSC and MCC, and the distance-based approach AHD since they are well-established and commonly used. 

\begin{table}[ht]
    \caption{List of systematic errors considered in our study.}\label{tab:SyntheticSegmentations}
    \begin{tabular}{>{\centering\arraybackslash}m{0.15\linewidth} >{\centering\arraybackslash}m{0.18\linewidth} p{0.5\linewidth}}\hline
        \textbf{Category} & \textbf{Label} & \textbf{Creation Process} \\\hline\hline
        \multirow{3}{=}{Proximate} & Erosion & morphological erosion with ball\\\cline{2-3}
         & Dilation & morphological dilation with ball \\\cline{2-3}
         & Fuzzy Edge & uniform in the band Dilation - Erosion \\\hline
        \multirow{2}{=}{Distant} 
        & FN Cluster & threshold EDT of GT foreground\\\cline{2-3}
        & FP Cluster & threshold EDT of GT background\\\hline
        \multirow{2}{=}{Random} & Uniform & uniform random errors \\\cline{2-3}
         & Nonuniform & inhomogeneous Poisson point process with vertically linearly decreasing intensity \\\hline
    \end{tabular}
\end{table}

\subsubsection{Traditional Metrics}
The metrics are illustrated for a Boolean model of cylinders with volume density 50\% and a system on non-overlapping cubes with volume density 10\% in Figures \ref{fig:MetricComparison} and \ref{fig:MetricComparison2}, respectively.
The traditional metrics DSC and MCC behave as one would expect from their definitions: Values are independent of the spatial distribution of errors and are influenced only by quantity and type. In particular, systematic errors which introduce the same amounts of FPs and FNs for a given structure produce the same value of DSC and MCC, e.g., dilation and FP cluster. This persists across all geometries and is expected from their respective definitions. In class-balanced settings as shown in Figure \ref{fig:MetricComparison}, both metrics correlate strongly with the error rate and do not provide additional characterization of quality. In cases of class imbalance, MCC and DSC give a higher priority to correctly labeling the less frequent foreground, see Figure \ref{fig:MetricComparison2}. This is usually desirable and the main advantage of these metrics over using the error rate. 

\subsubsection{Distance-based Metrics}\label{sec:DistanceMetrics}
\begin{figure*}[ht]
    \centering
    \begin{subfigure}{0.24\linewidth}
        \centering
        \includegraphics[width=0.9\linewidth]{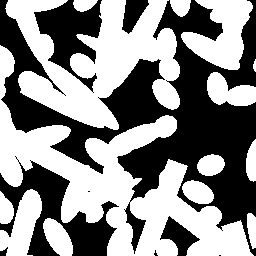}
        \subcaption{Section Image}
    \end{subfigure}
    \begin{subfigure}{0.24\linewidth}
        \includegraphics[width=\linewidth,trim={0 0 0.9cm 0.7cm},clip]{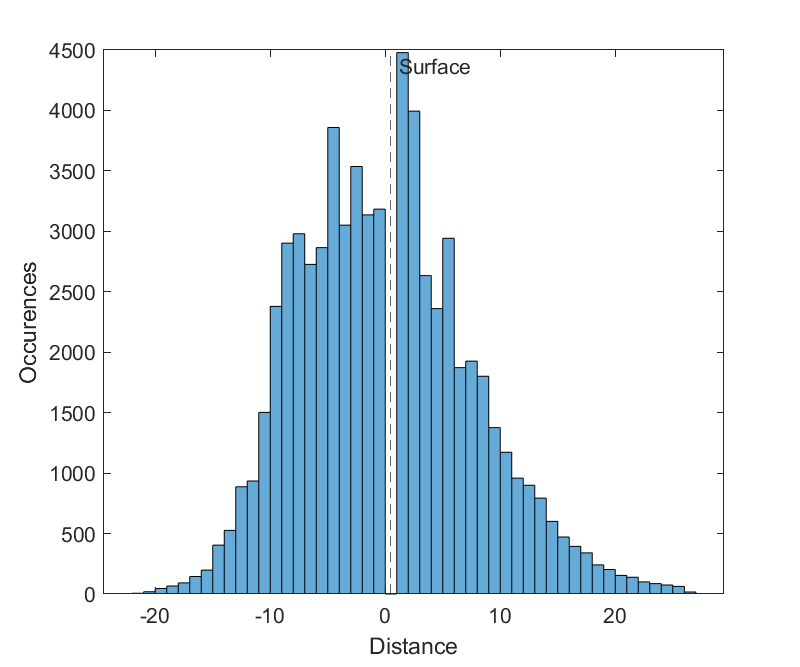}
        \subcaption{Distance Profile}
    \end{subfigure}
    \begin{subfigure}{0.24\linewidth}
        \includegraphics[width=\linewidth,trim={0 0 0.9cm 0.7cm},clip]{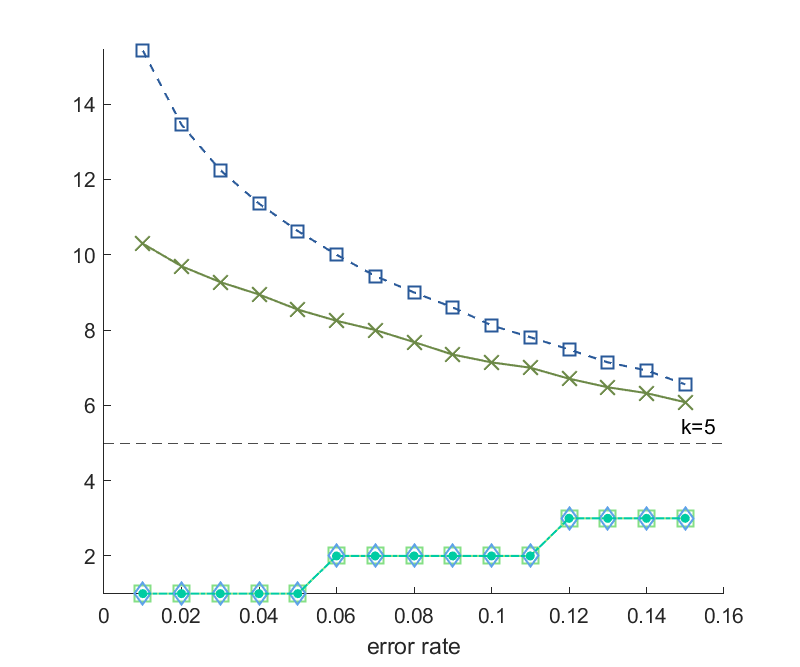}
        \subcaption{Extreme Distance}
    \end{subfigure}
    \begin{subfigure}{0.24\linewidth}
        \centering
        \includegraphics[width=0.8\linewidth]{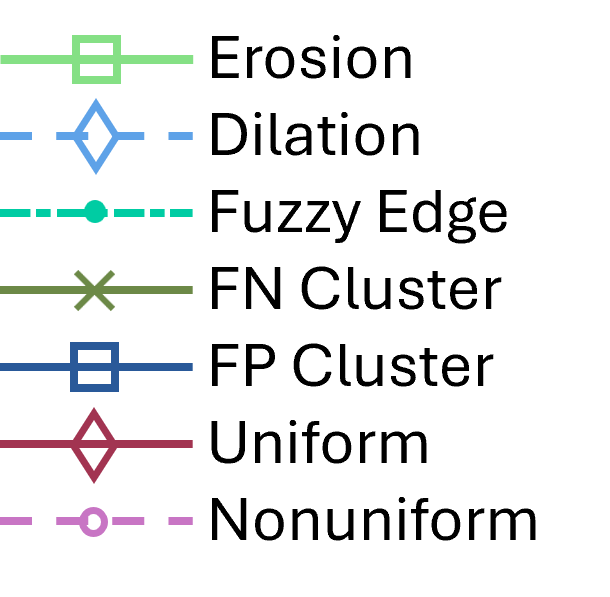}
        \subcaption{Legend}
    \end{subfigure}
    \begin{subfigure}{0.24\linewidth}
        \includegraphics[width=\linewidth,trim={0 0 0.9cm 0.7cm},clip]{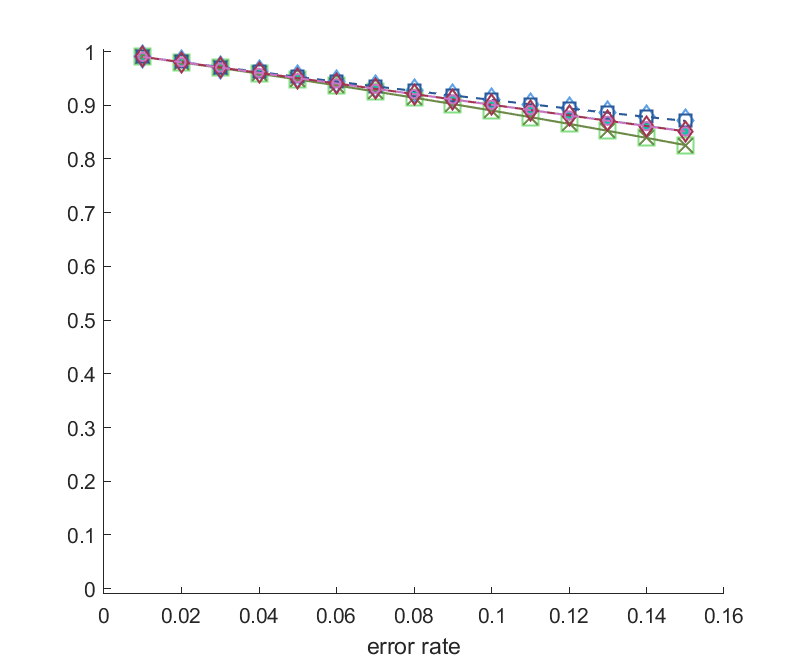}
        \subcaption{DSC}
    \end{subfigure}
    \begin{subfigure}{0.24\linewidth}
        \includegraphics[width=\linewidth,trim={0 0 0.9cm 0.7cm},clip]{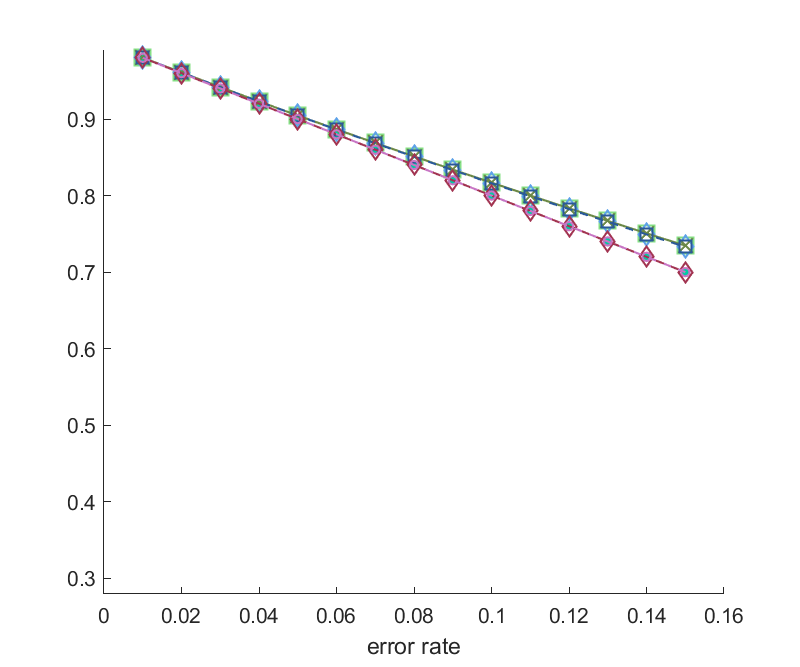}
        \subcaption{MCC}
    \end{subfigure}
    \begin{subfigure}{0.24\linewidth}
        \includegraphics[width=\linewidth,trim={0 0 0.9cm 0.7cm},clip]{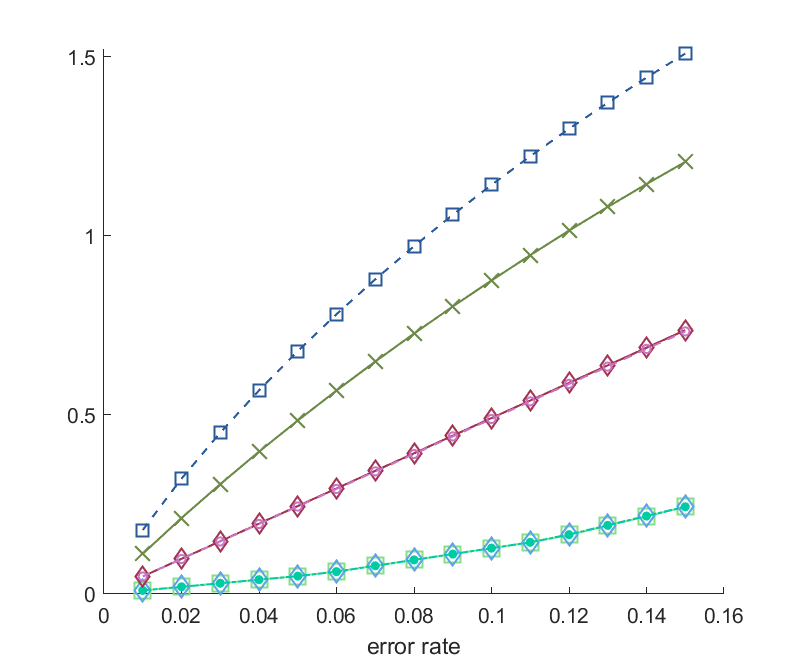}
        \subcaption{AHD}
    \end{subfigure}
    \begin{subfigure}{0.24\linewidth}
        \includegraphics[width=\linewidth,trim={0 0 0.9cm 0.7cm},clip]{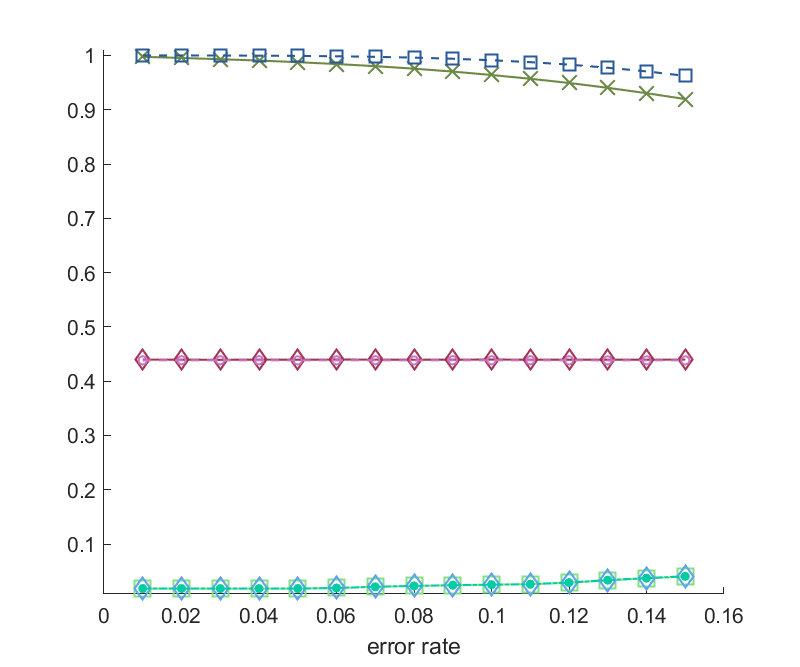}
        \subcaption{SCC$_{1,5}$}
    \end{subfigure}
    \caption{Comparison of quality metrics for systematic errors on an image of overlapping cylinders of radius 10.5 pixels, height 210 pixels and volume density 50\%. (a) sectional image of the structure. (b) histogram of distances from the surface split into foreground (negative) and background (positive) pixels. (c) largest distance of an error to the surface for proximate errors and smallest distance to the surface for distant errors. (e-f) Metrics for varying error rate.}
    \label{fig:MetricComparison}
\end{figure*}

\begin{figure*}[ht]
    \centering
    \begin{subfigure}{0.24\linewidth}
        \centering
        \includegraphics[width=0.9\linewidth]{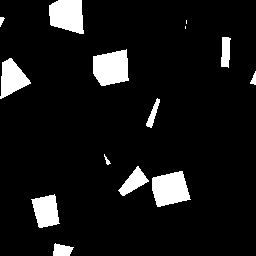}
        \subcaption{Section Image}
    \end{subfigure}
    \begin{subfigure}{0.24\linewidth}
        \includegraphics[width=\linewidth,trim={0 0 0.9cm 0.7cm},clip]{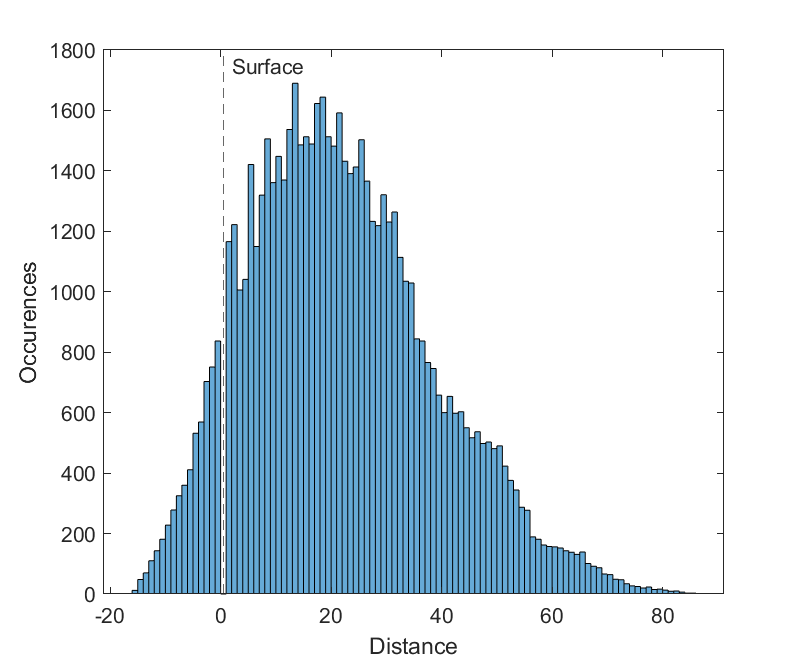}
        \subcaption{Distance Profile}
    \end{subfigure}
    \begin{subfigure}{0.24\linewidth}
        \includegraphics[width=\linewidth,trim={0 0 0.9cm 0.7cm},clip]{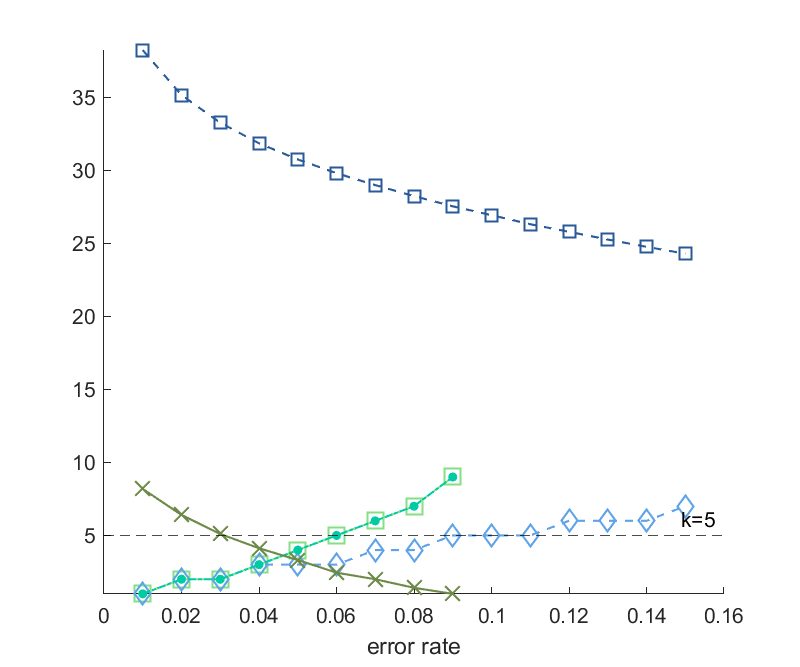}
        \subcaption{Extreme Distance}
    \end{subfigure}
    \begin{subfigure}{0.24\linewidth}
        \centering
        \includegraphics[width=0.8\linewidth]{Evaluation-Images/Legend.png}
        \subcaption{Legend}
    \end{subfigure}
    \begin{subfigure}{0.24\linewidth}
        \includegraphics[width=\linewidth,trim={0 0 0.9cm 0.7cm},clip]{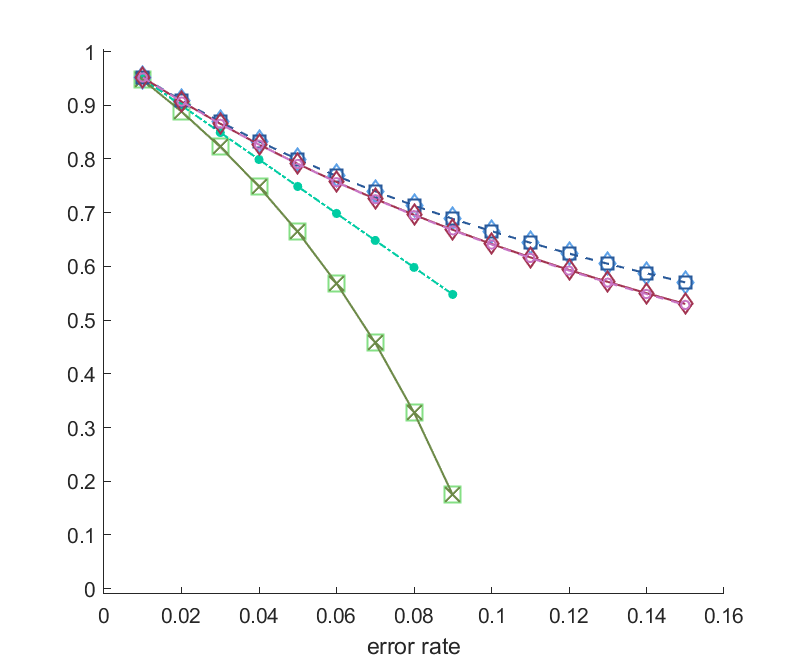}
        \subcaption{DSC}
    \end{subfigure}
    \begin{subfigure}{0.24\linewidth}
        \includegraphics[width=\linewidth,trim={0 0 0.9cm 0.7cm},clip]{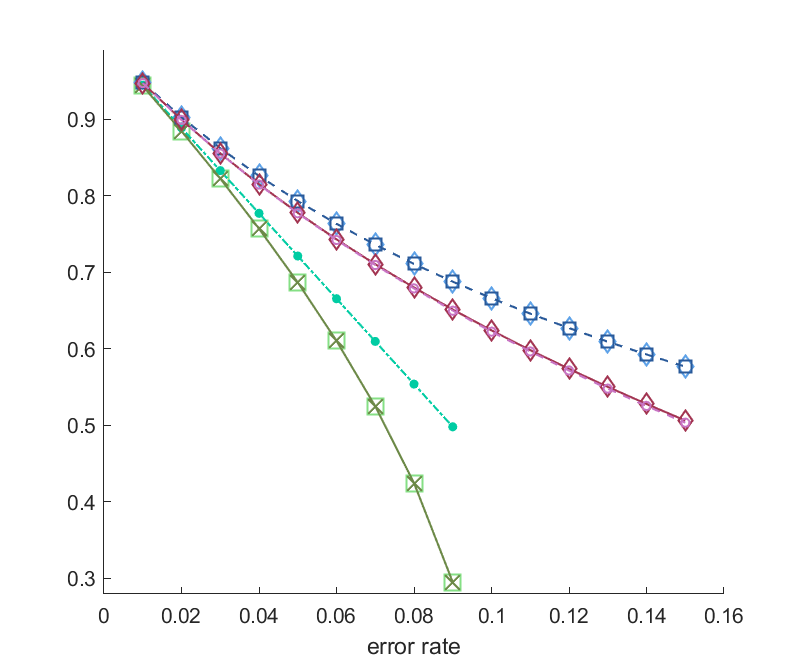}
        \subcaption{MCC}
    \end{subfigure}
    \begin{subfigure}{0.24\linewidth}
        \includegraphics[width=\linewidth,trim={0 0 0.9cm 0.7cm},clip]{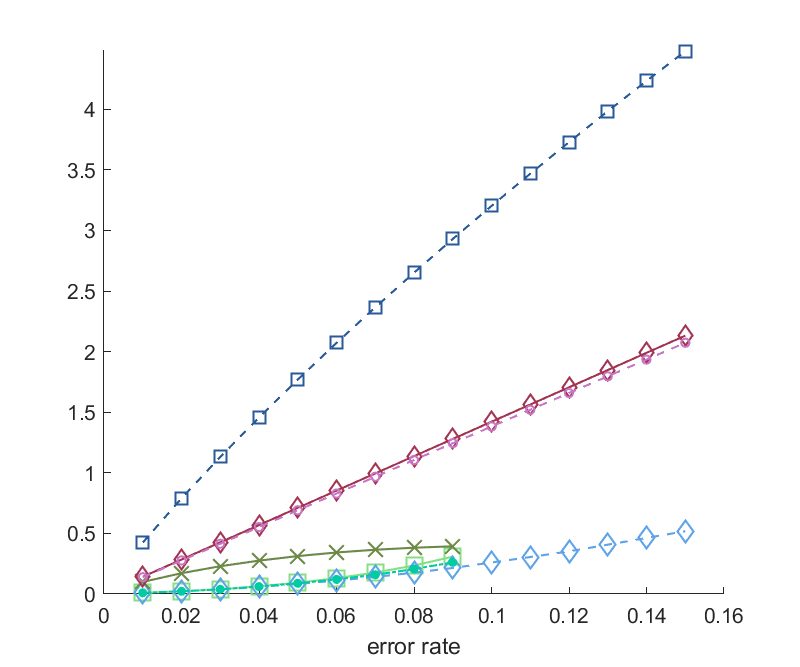}
        \subcaption{AHD}
    \end{subfigure}
    \begin{subfigure}{0.24\linewidth}
        \includegraphics[width=\linewidth,trim={0 0 0.9cm 0.7cm},clip]{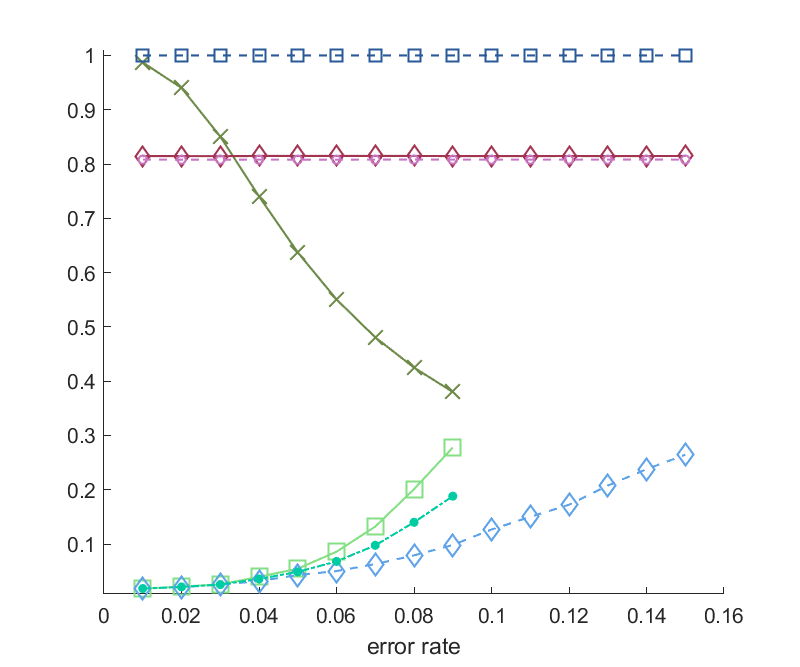}
        \subcaption{SCC$_{1,5}$}
    \end{subfigure}
    \caption{Comparison of quality metrics for systematic errors on an image of non-overlapping cubes with side length 30 pixels and volume density 10\%. (a) sectional image of the structure. (b) histogram of distances from the surface split into foreground (negative) and background (positive) pixels. (c) largest distance of an error to the surface for proximate errors and smallest distance to the surface for distant errors. (e-f) Metrics for varying error rate.}
    \label{fig:MetricComparison2}
\end{figure*}

The distance-based metrics AHD and SCC behave very differently on the systematic errors. They are able to separate proximate from distant errors and are not influenced by the error type. AHD achieves this by incorporating both the quantity of errors and their distance from the surface yielding almost linear scaling with the error rate. The scaling for distant errors is distinctively larger than for proximate errors, see Figure \ref{fig:MetricComparison} (g). However, individual values of AHD are difficult to interpret and compare. This has two reasons. First, the same value of AHD can be obtained through different combinations of error quantity and their positions, e.g., few distant or many proximal errors. Second, there exists no uniform supremum between geometries such that a deduction of quality is not possible from a specific value, compare Figures \ref{fig:MetricComparison} and \ref{fig:MetricComparison2} (g). Nevertheless, when combining AHD with traditional metrics comparative statements about quality differences based on the spatial distribution of errors can be made when comparing different segmentations. However, this requires considerable expertise and effort since the influence of different influential factors must be recognized.

\begin{figure*}[ht]
    \centering
    \begin{subfigure}{0.24\linewidth}
        \centering
        \includegraphics[width=0.9\linewidth]{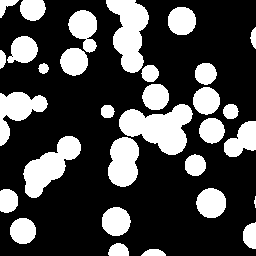}
        \subcaption{Section Image}
    \end{subfigure}
    \begin{subfigure}{0.24\linewidth}
        \includegraphics[width=\linewidth,trim={0 0 0.9cm 0.7cm},clip]{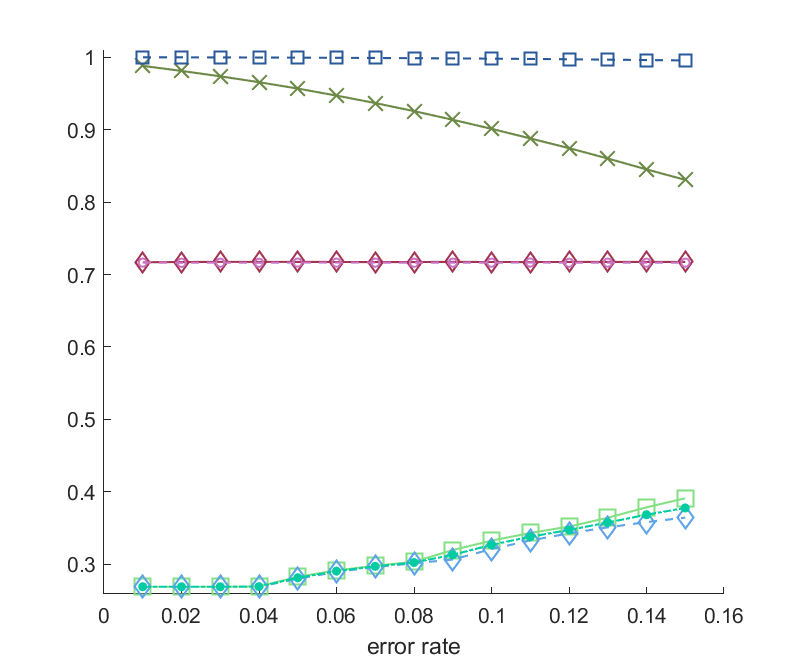}
        \subcaption{SCC$_{0.5,3}$}
    \end{subfigure}
    \begin{subfigure}{0.24\linewidth}
        \includegraphics[width=\linewidth,trim={0 0 0.9cm 0.7cm},clip]{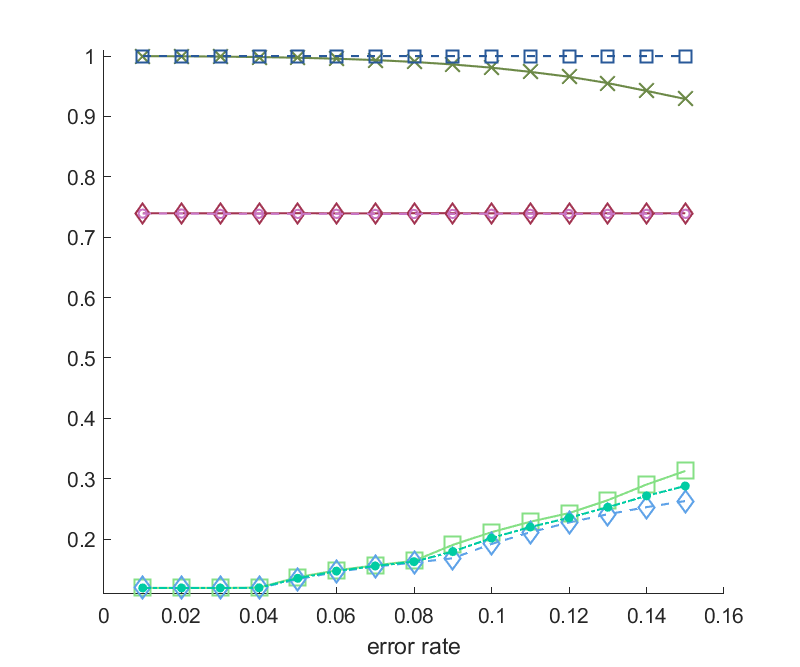}
        \subcaption{SCC$_{1,3}$}
    \end{subfigure}
    \begin{subfigure}{0.24\linewidth}
        \includegraphics[width=\linewidth,trim={0 0 0.9cm 0.7cm},clip]{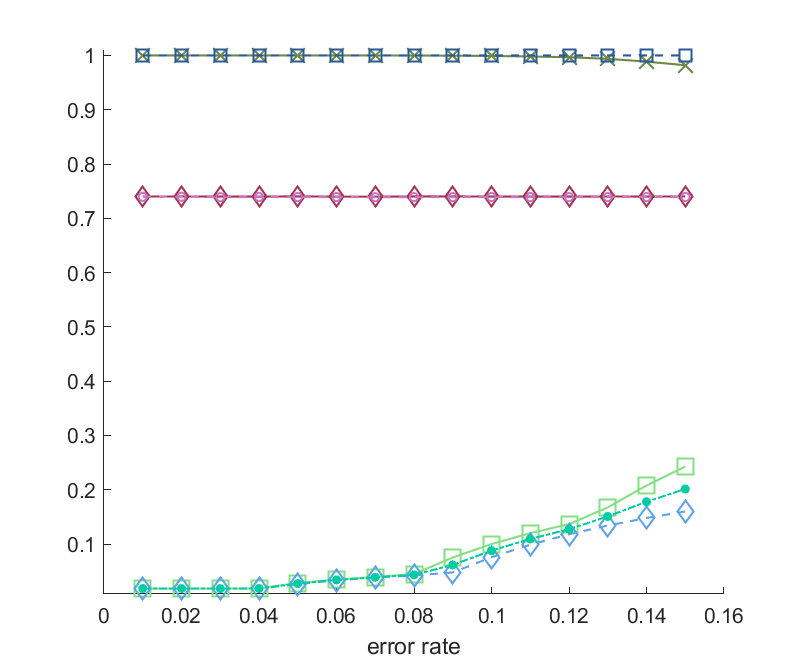}
        \subcaption{SCC$_{2,3}$}
    \end{subfigure}
    \begin{subfigure}{0.24\linewidth}
        \includegraphics[width=\linewidth,trim={0 0 0.9cm 0.7cm},clip]{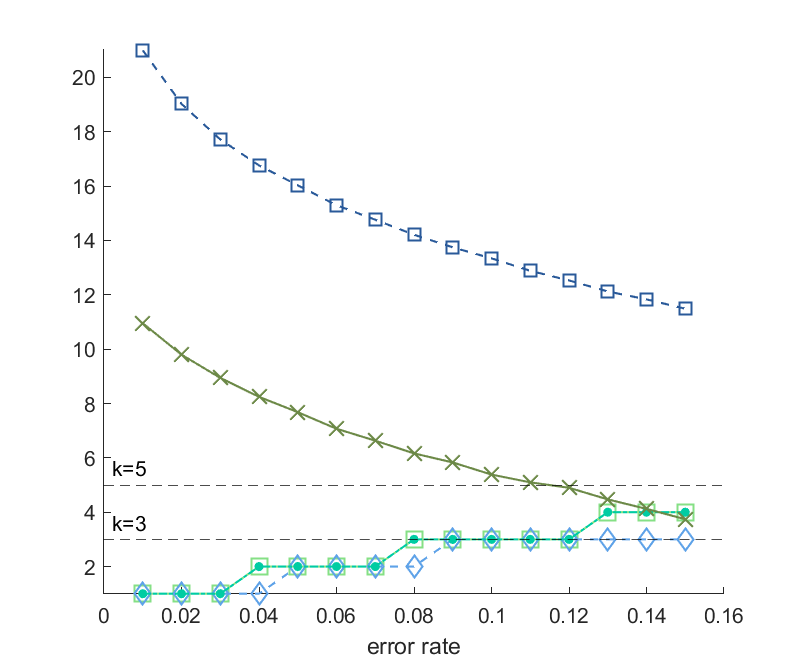}
        \subcaption{Extreme Distance}
    \end{subfigure}
    \begin{subfigure}{0.24\linewidth}
        \includegraphics[width=\linewidth,trim={0 0 0.9cm 0.7cm},clip]{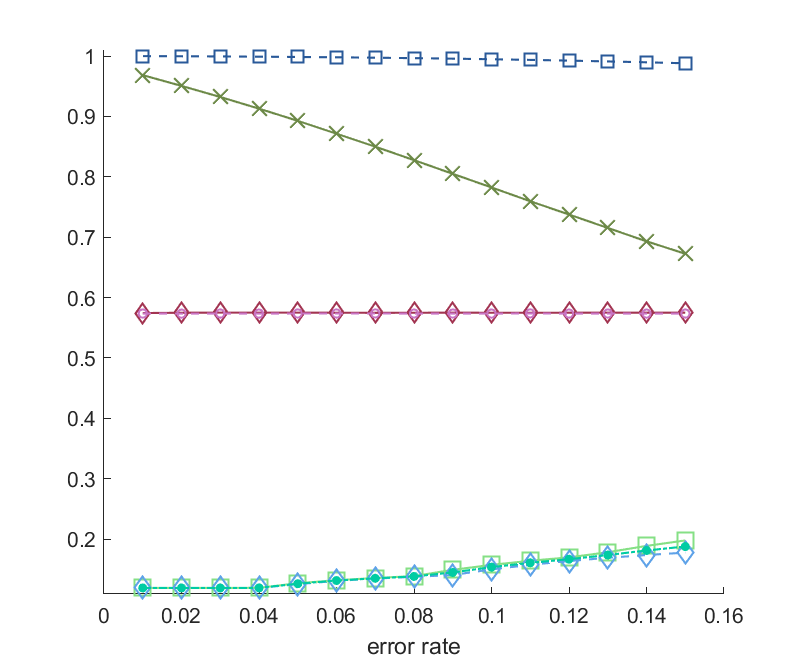}
        \subcaption{SCC$_{0.5,5}$}
    \end{subfigure}
    \begin{subfigure}{0.24\linewidth}
        \includegraphics[width=\linewidth,trim={0 0 0.9cm 0.7cm},clip]{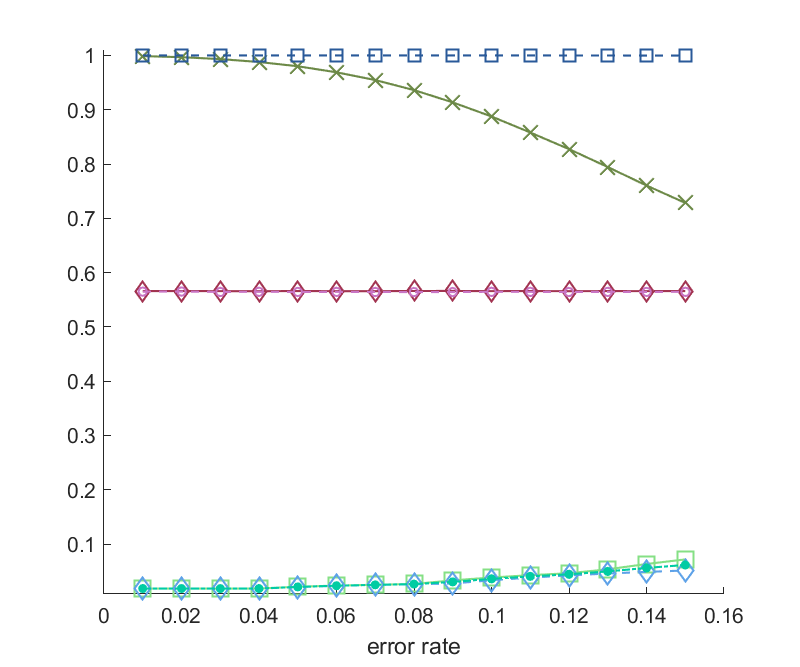}
        \subcaption{SCC$_{1,5}$}
    \end{subfigure}
    \begin{subfigure}{0.24\linewidth}
        \includegraphics[width=\linewidth,trim={0 0 0.9cm 0.7cm},clip]{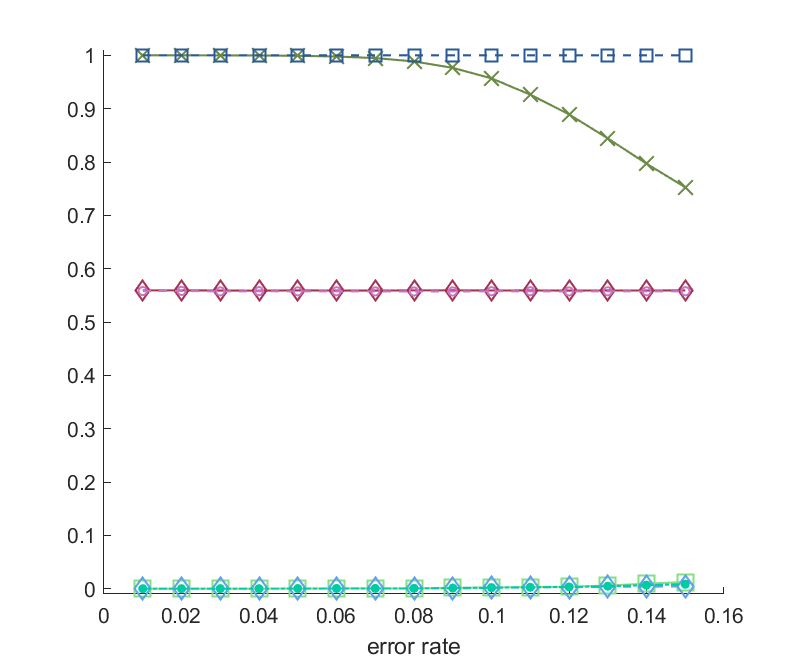}
        \subcaption{SCC$_{2,5}$}
    \end{subfigure}
    \caption{Evaluation of SCC on an image of overlapping spheres with radius 20 and volume density 30\% (a). (b)-(d) and (f)-(g) results for varying parameters $a$ and $k$. (e) Largest distance of an error to the surface for proximate errors and the smallest distance for distant errors. See Figure \ref{fig:AllGeometries} (e) for legend. }
    \label{fig:SCCParameters}
\end{figure*}

\begin{figure*}[ht]
    \centering
    \begin{subfigure}{0.24\linewidth}
		\centering
		\includegraphics[width=\linewidth,trim={0 0 0.9cm 0.7cm},clip]{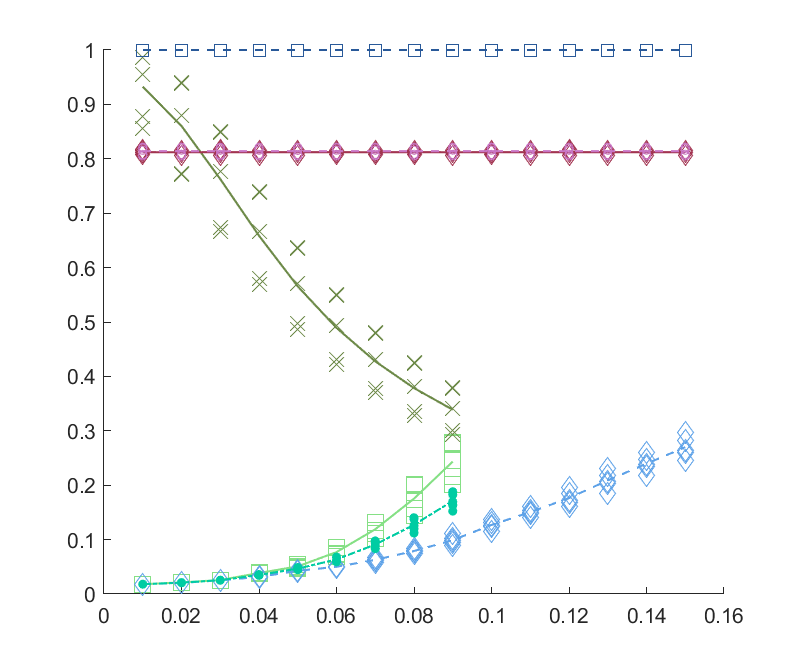}
		\subcaption{Density 10\%}
	\end{subfigure}
	\begin{subfigure}{0.24\linewidth}
		\includegraphics[width=\linewidth,trim={0 0 0.9cm 0.7cm},clip]{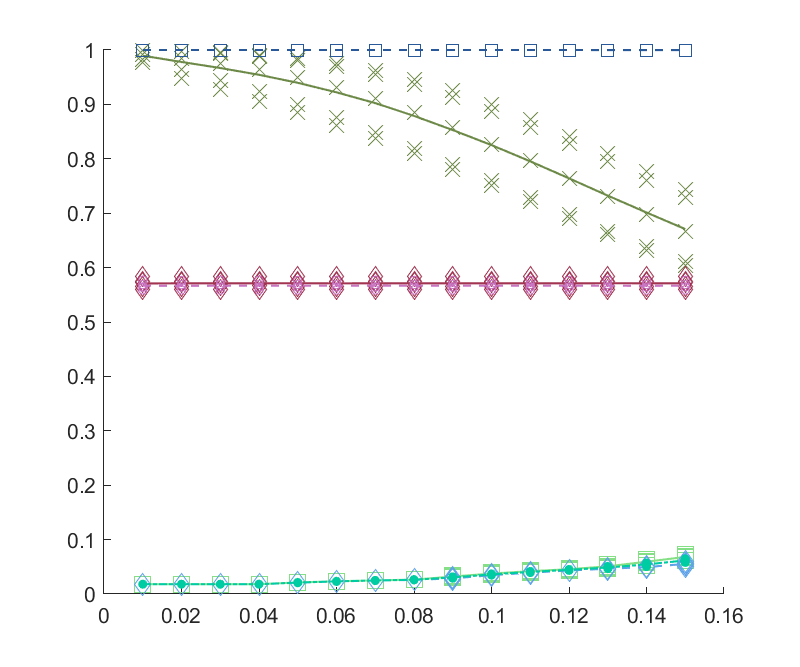}
		\subcaption{Density 30\%}
	\end{subfigure}
	\begin{subfigure}{0.24\linewidth}
		\includegraphics[width=\linewidth,trim={0 0 0.9cm 0.7cm},clip]{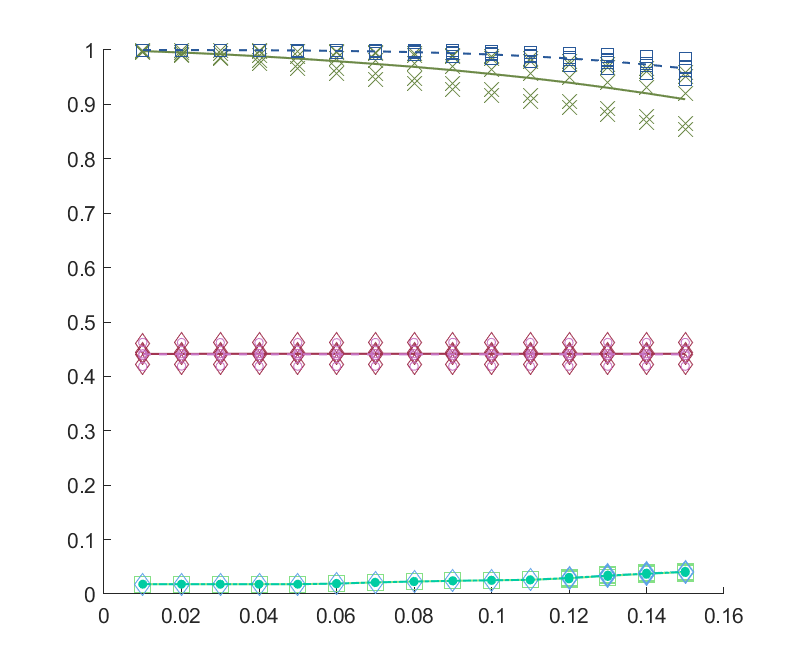}
		\subcaption{Density 50\%}
	\end{subfigure}
	\begin{subfigure}{0.24\linewidth}
		\includegraphics[width=\linewidth,trim={0 0 0.9cm 0.7cm},clip]{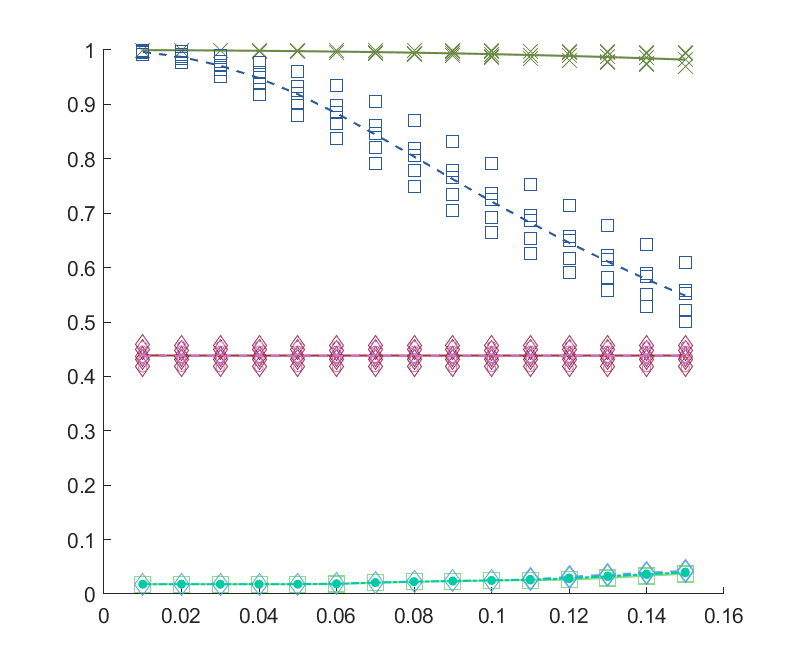}
		\subcaption{Density 70\%}
	\end{subfigure}
    \caption{Evaluation of SCC on the entire data set split by volume density. Lines show the corresponding mean for each systematic error. See Figure \ref{fig:MetricComparison2} (d) for legend.}
    \label{fig:AllGeometries}
\end{figure*}

The problems observed with AHD are entirely resolved by SCC. Most notably, our quality metric is easily interpretable and uniquely characterizes error positioning without the influence of error type or quantity, at least in the balanced design. Proximal errors are consistently mapped to zero and distant errors to one, see Figure \ref{fig:MetricComparison} (h). Depending on the application and preference of the user, either extreme may reflect better quality. In general, the value of SCC relates to the fraction of errors found outside the proximity range. In case of random predictions which place incorrect labels independently of the structure, this translates to the fraction of volume outside the proximity range, which is 43\% for the structure in Figure \ref{fig:MetricComparison} and 82\% in Figure \ref{fig:MetricComparison2} and is in line with our measurements. This interpretation is further substantiated when we consider the behavior for larger error rates in Figure \ref{fig:MetricComparison2} (h). Due to the low volume density, the distance profile is so that larger error rates of the proximate errors and the FN cluster are achieved only by introducing incorrect labels violating the proximity range, see Figure \ref{fig:MetricComparison2} (b-c). In other words, the proximate errors introduce incorrect labels which are not considered proximate, and similarly, FN cluster introduces incorrect labels which are not distant anymore. The same argument applies when considering different parameter combinations of $a$ and $k$ as shown in Figure \ref{fig:SCCParameters}. Here, the structure and distance distribution remain the same but the definition of proximity changes. Deviations from the anticipated values for proximate and distant errors again relate directly to violations of the respective proximity range. This demonstrates that SCC can comprehensively quantify the spatial distribution of errors regardless of the error rate and type. In addition to this, SCC yields consistent values for similar spatial distributions of errors. This is shown in Figure \ref{fig:AllGeometries} where variations in the SCC remain small across comparable geometries, i.e. geometries with comparable distance profiles and therefore error distributions.

We propose pairing SCC with a traditional quality metric encoding the error rate. In this way, both the number of erroneous pixels and their spatial locations can be incorporated into the validation of a segmentation result.

\begin{figure*}[ht]
    \centering
    \begin{subfigure}{0.24\linewidth}
        \includegraphics[width=\linewidth,trim={2cm 2cm 2cm 2cm},clip]{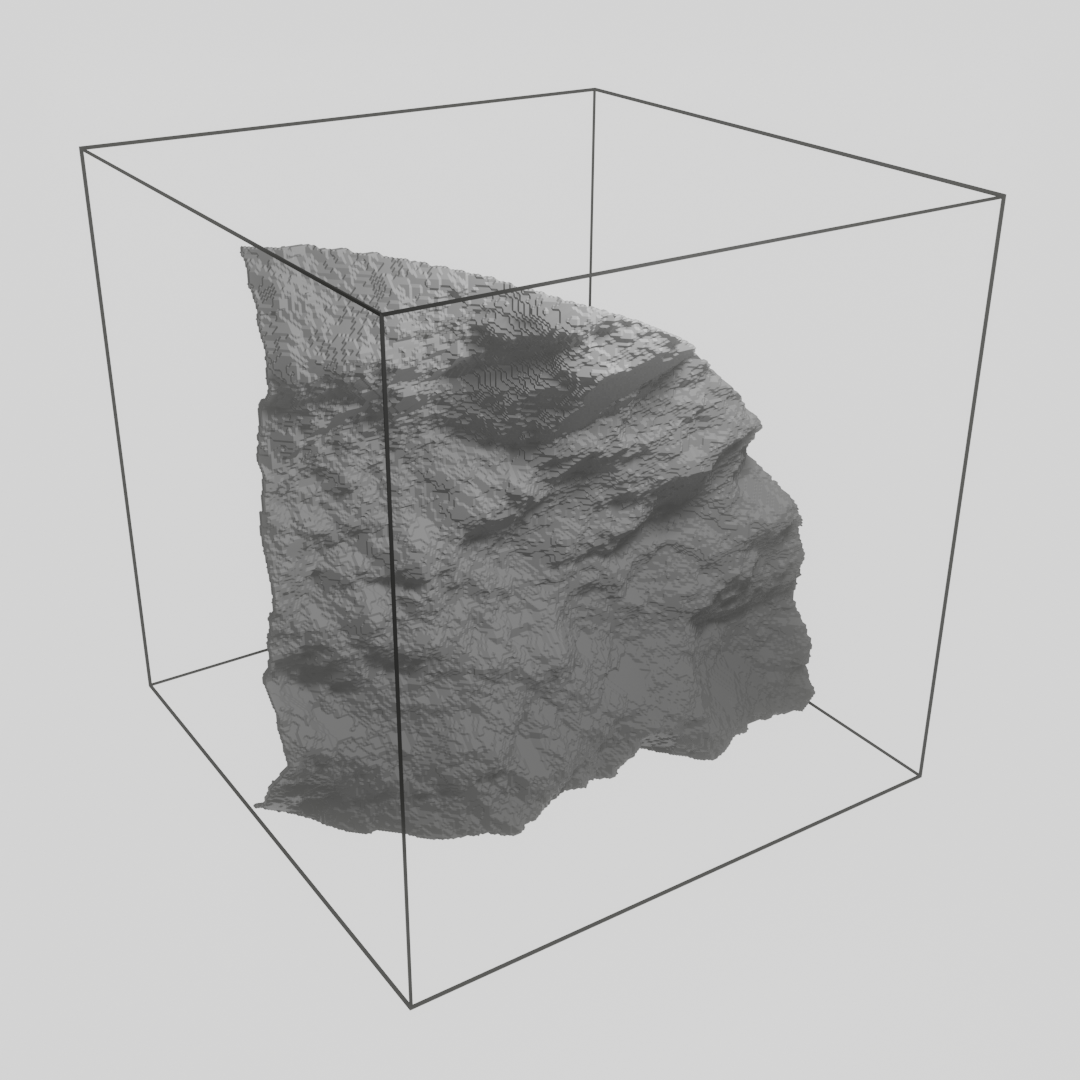}
        \subcaption{Ground Truth}
    \end{subfigure}
    \begin{subfigure}{0.24\linewidth}
        \includegraphics[width=\linewidth,trim={2cm 2cm 2cm 2cm},clip]{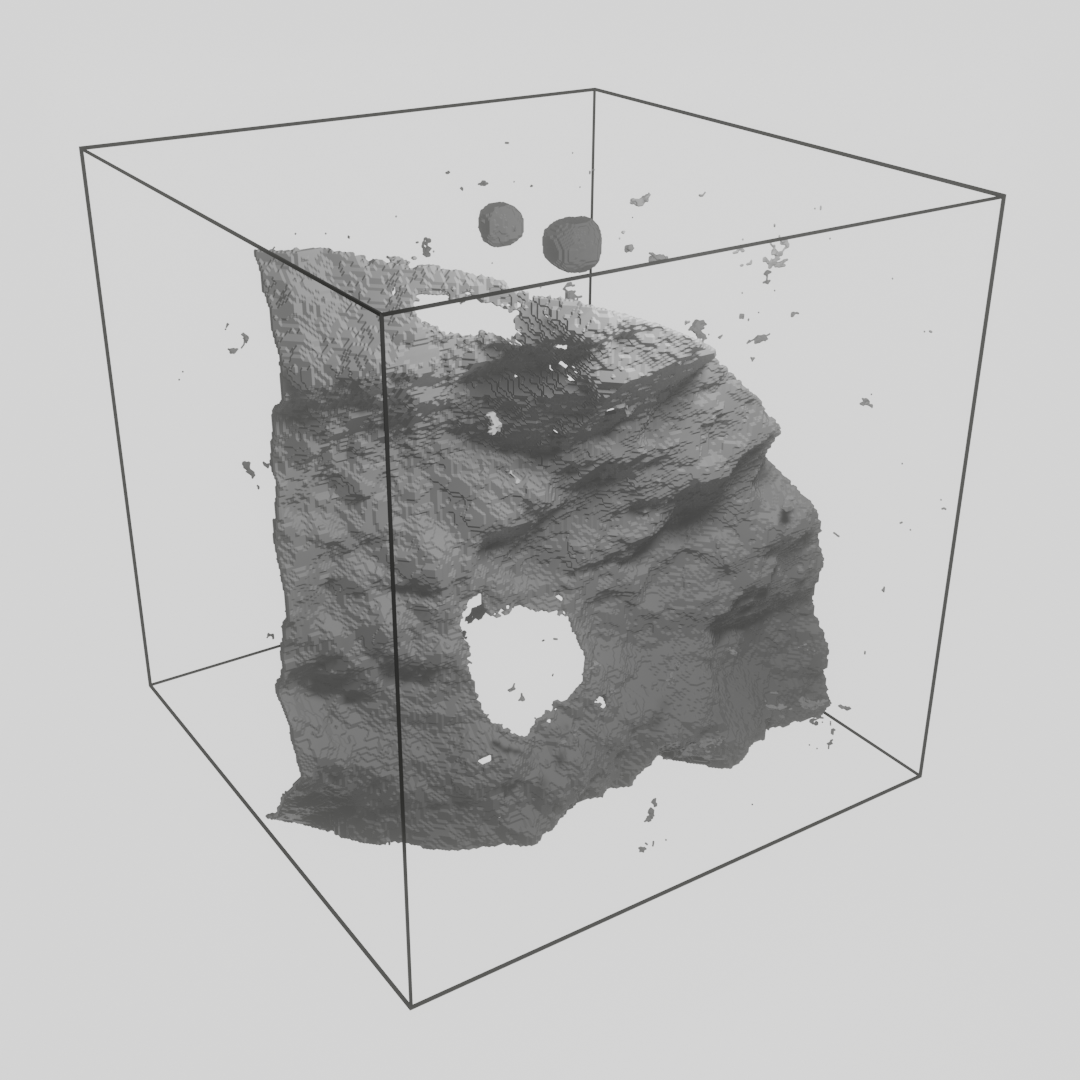}
        \subcaption{Percolation}
    \end{subfigure}
    \begin{subfigure}{0.24\linewidth}
        \includegraphics[width=\linewidth,trim={2cm 2cm 2cm 2cm},clip]{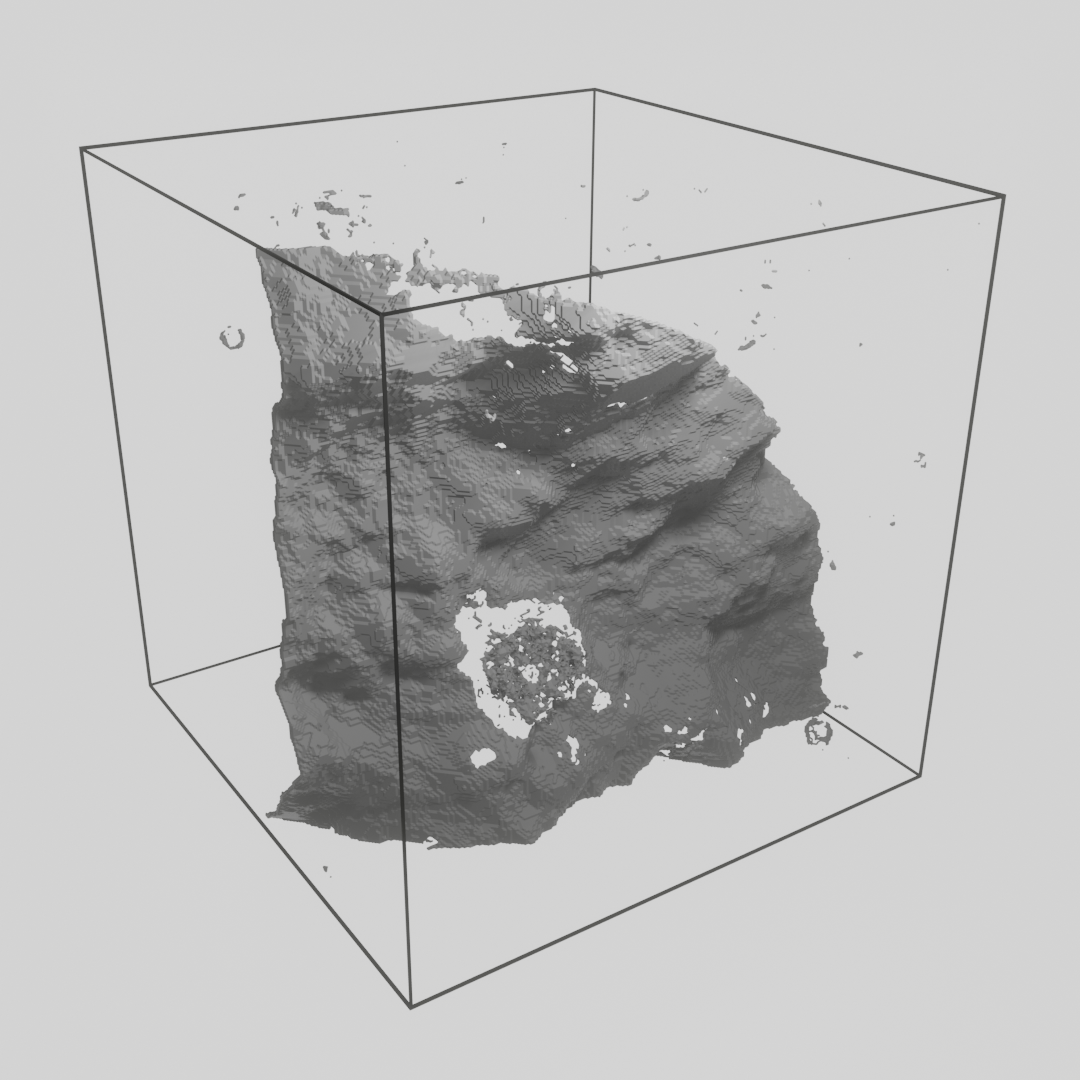}
        \subcaption{Random Forest}
    \end{subfigure}
    \begin{subfigure}{0.24\linewidth}
        \includegraphics[width=\linewidth,trim={2cm 2cm 2cm 2cm},clip]{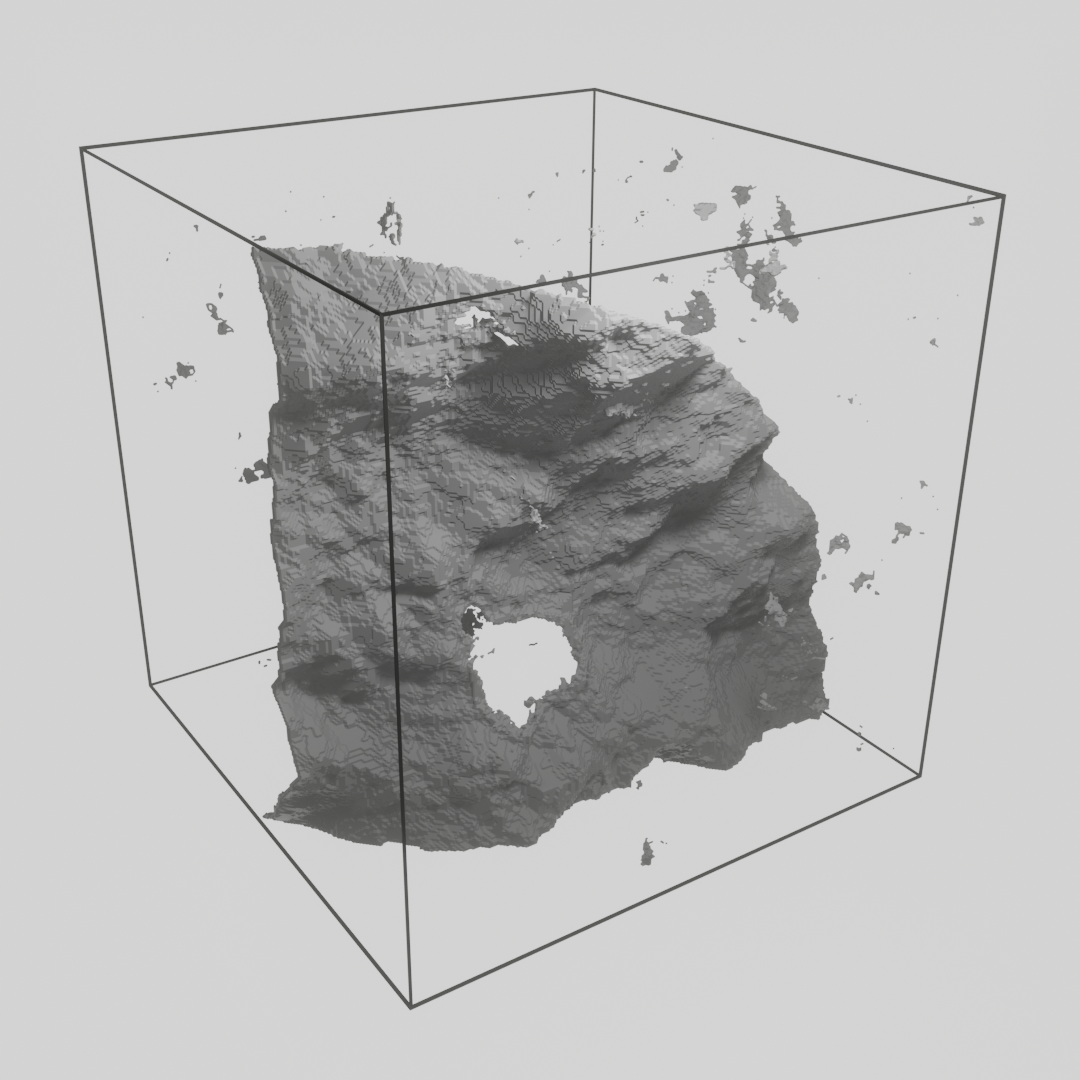}
        \subcaption{3D U-Net}
    \end{subfigure}
    \caption{(a) Renderings of synthetic crack image of width 3 pixels and (b-d) segmentation results obtained from different methods \cite{barisin_methods_2022}.}
    \label{fig:CrackResults}
\end{figure*}

\subsection{Application: Cracks in Concrete}
In the following, we show how SCC can be used to extend and improve the quality assessment of traditional metrics. Therefore, we consider exemplary data for a synthetic crack in concrete that was generated and initially evaluated by Barisin et al. \cite{barisin_methods_2022}. We refer to \cite{jung_crack_2023} for further details on the crack simulation and \cite{jung_vorocrack3d_2024} for a published data set. The data have a high class imbalance and consist of a ground truth geometry of a thin crack and corresponding annotations which were generated by different segmentation methods, see Figure \ref{fig:CrackResults}. In their article, the group found that learning-based methods, random forest and 3D U-net, and Hessian-based percolation produce the best results of all tested methods by using visual assessment and traditional metrics. In their quality assessment, they also considered including a tolerance to ignore incorrect labels adjacent to the crack. This is not unusual for segmentation tasks and follows the idea that capturing the general shape of the crack rather than its precise texture is often more important. SCC incorporates this concept directly without ignoring errors, since it yields small values when errors are predominantly located near the surface, i.e., when the general shape is met. 

\begin{table}[ht]
    \centering
    \begin{tabular}{|c|c|c|c|c|c|}\hline
         Method & Error [\%] & DSC  & MCC & AHD & SCC$_{2,3}$ \\\hline\hline
         Percolation & 0.26 & 0.868 & 0.868 & 0.094 & 0.317 \\\hline
         Random Forest & 0.23 & 0.866 & 0.866 & 0.007 & 0.080 \\\hline
         3D U-Net & 0.21 & 0.892 & 0.893 & 0.026 & 0.177 \\\hline
    \end{tabular}
    \caption{Numerical assessment of segmentation results shown in Figure \ref{fig:CrackResults}.}
    \label{tab:ConcreteValues}
\end{table}

Table \ref{tab:ConcreteValues} displays the numerical assessment of quality without tolerance using the error rate, DSC, MCC, AHD and SCC on the three methods which were found to perform well. In this example, we use a different definition of proximity with $a=2$ and $k=3$ when calculating the SCC due to the small thickness of the crack. Based only on traditional metrics, we agree with the authors that the three methods perform similarly well and that 3D U-net is marginally better. However, by including SCC significant differences in the spatial distribution of errors are revealed. In particular, the lower value for random forest indicates that it captures the general shape better than the 3D U-net since errors are more concentrated near the surface. However, the larger value for percolation implies that the surface texture is more precisely reproduced, and errors are mostly introduced further from the surface, e.g., when falsely detecting pores or pore boundaries. In such cases, post-processing, such as the removal of smaller connected components, can be employed.

We want to highlight that with the help of SCC we were able to improve on previous quality assessment and determine that the result by 3D U-net is not universally optimal. Depending on the user's goal to accurately capture the general shape or precise surface texture, either random forest or percolation may be the method of choice. In the crack segmentation example, arguments can be made for both sides. Detecting the general areas where cracks are formed is essential for analyzing the structural integrity and mechanical failure of concrete. However, accurate surface texture is required to understand, model, and predict crack formation.

\section{Conclusion}
In this work, we have proposed, analyzed, and validated a novel metric called Surface Consistency Coefficient (SCC) to assess quality in image segmentation based on geometric relations. Compared to established distance-based metrics, SCC is normalized and yields consistent values summarizing the distribution of errors in terms of distance from the surface. This makes it easy to interpret and compare for different methods and datasets. Additionally, SCC remains independent of the quantity of errors, allowing it to be combined with traditional metrics such as accuracy or error rate to quantify the number of errors and their geometric distribution independently. This reduces redundant information found in other distance-based quality metrics and enables a more precise assessment of quality and comparison of segmentations.


\bmsection*{Acknowledgments}
The data of sythetic cracks was created as part of the BMBF project DAnoBi (Funding number: 05M2020). We thank Christian Jung for providing the corresponding renderings in Fig. \ref{fig:CrackResults} and insight into the data itself. 

\bmsection*{Financial disclosure}
Federal Ministry of Education and Research (BMBF), Project: Synthetic Data for Machine Learning Segmentation of Highly Porous Structures from FIB-SEM Nano-tomographic Data (poSt), Funding number: 01IS21054A

\bmsection*{Conflict of interest}
The authors declare no potential conflict of interests.

\bibliography{Segmentierungsmasse}

\end{document}